# Engineering a Simplified 0-Bit Consistent Weighted Sampling

Edward Raff
Booz Allen Hamilton
raff_edward@bah.com

Jared Sylvester
Booz Allen Hamilton
sylvester_jared@bah.com

Charles Nicholas
Univ. of Maryland, Baltimore County
nicholas@umbc.edu

## ABSTRACT

The Min-Hashing approach to sketching has become an important tool in data analysis, information retrial, and classification. To apply it to real-valued datasets, the ICWS algorithm has become a seminal approach that is widely used, and provides state-of-the-art performance for this problem space. However, ICWS suffers a computational burden as the sketch size $K$ increases. We develop a new Simplified approach to the ICWS algorithm, that enables us to obtain over 20x speedups compared to the standard algorithm. The veracity of our approach is demonstrated empirically on multiple datasets and scenarios, showing that our new Simplified CWS obtains the same quality of results while being an order of magnitude faster.

## 1 INTRODUCTION

The well known Jaccard similarity provides a valid kernel for measuring the similarity between sets. Given one set $S$ and a second set $O$, it simply returns the ratio of their intersection over their union, $J(S, O) = \frac{|S \cap O|}{|S \cup O|}$. Seminal work by Broder introduced the min-hashing idea, allowing $J(S, O)$ to be computed accurately and efficiently by keeping only sketches of each set $S$ and $O$, where a sketch is a sub-set of the original sets [1, 2, 5]. Min-Hashing has been used for effective and fast personalized recommendations algorithms [4], near-duplicate detection for web-pages [14] and images [3], malware clustering and classification [9, 16], and general information retrieval problems [20].

Given a min-hash function *Minhash*, we can increase or decrease the sketch size $K$ to increase accuracy of the approximation or decrease the storage cost and compute time of the sketch. Algorithm 1 demonstrates its operation, and is used by all min-hashing algorithms. This works because the probability of two sets producing the same min-hash for a given seed ($k$) is equal to the Jaccard similarity itself (i.e., $\forall k, P(\text{Minhash}(S, k) = \text{Minhash}(O, k)) = J(S, O)$). Min-hashing can thus be seen as a sampling method to compute the similarity. Since the required min-hashes can be computed once per set, and require only an equality check, they are often faster to use in practice, especially for large systems.

**Algorithm 1** MinHash Approximation
---
**Require:** Two sets $S$ and $O$ that we want to compute the similarity of.
1: $s \leftarrow 0$
2: **for** $k \in 1, 2, \ldots, K$ **do**
3:     **if** $\text{Minhash}(S, k) = \text{Minhash}(O, k)$ **then**
4:         $s \leftarrow s + 1$
5:     **end if**
6: **end for**
7: **return** $s/K$

---

In this work, we are interested in applications in which items from the set are weighted. That is to say, given an entry $z \in S$, we associate with $z$ a positive value given by $w(S, z)$. We can then say that $\forall z \in S, w(S, z) > 0$. If a value $q \notin S$, then $w(S, q) = 0$. The Weighted Jaccard Similarity (WJS) (1), also known as the min-max kernel, is the generalization of the $J(S, O)$ to this use case, and $WJS(S, O) = J(S, O)$ when all weights are equal. While computing min-hashes for approximating the Jaccard similarity can be done in time $O(D)$, where $D$ is the number of items in the set [1, 2], approximating the WJS requires a more expensive $O(DK)$ time per-set. Reducing the compute time required to construct these hashes is the focus of this work. We obtain constant time speedups of over 20x by mathematically simplifying the current approaches to sketching the WJS and then exploiting this simplicity to produce a simple and compact approximation.

$$WJS(S, O) = \frac{\sum_{\forall z \in S \cup O} \min(w(S, z), w(O, z))}{\sum_{\forall z \in S \cup O} \max(w(S, z), w(O, z)))} \quad (1)$$

Manasse et al. [13] proposed the Consistent Weighted Sampling (CWS) algorithm for the WJS problem. CWS produces a sketch of $K$ hashes directly from the weighted samples in the set. Each sample in the sketch has a probability of collision with a sample from another set equal to the WJS, which allows the WJS's estimation by taking multiple samples. This resulted in an ametorized $O^*(DK)$ time algorithm.

CWS was improved upon by Ioffe [8] to produce the Improved CWS (ICWS), which required only fixed constant time per hash, so producing $K$ hashes with a feature set of $D$ features takes $O(KD)$ time per data point. ICWS is considered the state of the art for approximating the WJS, as well as the $L_1$ distance [8]. The ICWS algorithm is presented in Algorithm 2.

The ICWS algorithms iterates through every item $z$ in the set $S$, and computes a value $a_z$ for each feature. The minimum $a_z$ determines the min-hash, and returns it's min-hash as a tuple of two values. Each value must be the same to count as a match. The value of $a_z$ is stochastic, which is necessary that different entries $z \in S$ will be selected for different hash indexes $k \in K$. To ensure that two different sets $S$ and $O$ select the same values when equal, the Pseudo Random Number Generator (PRNG) is seeded using the feature $z$ and hash index $k$.

**Algorithm 2** ICWS

1: **procedure** MINHASH(Weighted Set $S$, hash index $k$)
2: 　　**for all** $z \in S$ **do**
3: 　　　　Seed PRNG with tuple $(z, k)$
4: 　　　　$r_z \sim \text{Gamma}(2, 1)$
5: 　　　　$c_z \sim \text{Gamma}(2, 1)$
6: 　　　　$\beta_z \sim \text{Uniform}(0, 1)$
7: 　　　　$t_z \leftarrow \left\lfloor \frac{\log w(S, z)}{r_z} + \beta_z \right\rfloor$
8: 　　　　$y_z \leftarrow \exp(r_z(t_z - \beta_z))$
9: 　　　　$a_z \leftarrow \frac{c_z}{y_z \exp(r_z)}$
10: 　　**end for**
11: 　　$z^* \leftarrow \arg\min_z a_z$
12: 　　$y^* \leftarrow y_{z^*}$
13: 　　**return** tuple $z^*, t_{z^*}$
14: **end procedure**

While effective, the ICWS algorithm's $O(KD)$ cost can make it prohibitively expensive when large hash sizes $K$ are necessary, or when then the number of features $D$ is large. For most applications, values of $K$ in the hundreds or thousands is routinely necessary [19], and large values of $D$ are common for the information retrieval problems ICWS is often applied to [2, 18]. For this reason a number of works have looked at improving the runtime efficiency of the ICWS algorithm. Assuming all values are stored in 32-bit floats and integers,[1] the ICWS algorithm requires $2K$ memory for the sketch, and five random values sampled from the uniform distribution. This value of 5 comes from the two Gamma distributed values $r_z$ and $c_z$, which are computed as $x = -\log u_1 - \log u_2$, where $u_1, u_2 \sim \text{Uniform}(0, 1)$. Thus five uniform-random numbers need to be generated at each step, which has a significant cost [21]. In addition to these memory and sampling requirements, a non-trivial amount of expensive floating point operations are necessary. Lines 3 through 9 of Algorithm 2 requires five logarithms, two exponentiations, and four multiplications/divisions. We count all of these operations as they tend to be the most expensive to perform, and often overshadow the cost of floating point additions/subtractions or basic integer arithmetic.

In this work, we will introduce a simplified variant of the ICWS algorithm that improves hash creation time by an order of magnitude.

Our Simplified CWS strategy continues to work in all scenarios that ICWS does, such as when not all features are known in advance, or there is no cap on potential feature value magnitude. Further, existing techniques to improve ICWS's runtime are compatible with our approach.

Our improved runtime comes from reducing the work per feature and hash (lines 3-9) down to *just one floating-point multiplication*. This simplification is inspired by prior work that allows us to return a sketch requiring only $K$ memory units, which we will review with other related work in section 2. We will then derive our new Simplified CWS algorithm in section 3, and provide extensive empirical evidence of its quality and efficiency in section 4. Finally, we will conclude in section 5.

## 2 RELATED WORK

The original Confidence Weighted Sampling algorithm was introduced by Manasse et al. [13], providing the first direct sketching method for the Weighted Jaccard Similarity. Earlier work required reducing the WJS problem to the standard un-weighted Jaccard, but this results in an explosion in the feature set size and is unwieldy in practice [7]. The CWS algorithm was quickly improved upon by Ioffe [8], denoted as the seminal ICWS algorithm. This approach works with arbitrary non-negative weighted sets as inputs and requires no communication between points (i.e., every datum can be hashed independently of any other information). However, the computational burden of this approach is still non-trivial. For this reason Ioffe also proposed to reduce the set of each input to only the 200 most frequently selected features. This approach was shown to work well heuristically and gain a speedup of 150x. However this approach only allows a speedup at inference time, as the dataset must first be hashed (or "sketched") to determine the most frequently selected features. This necessarily re-introduces communication costs. In addition, the number of most frequently selected features that need to be kept will be problem dependent. We note that this inference speedup strategy is compatible with most ICWS extensions, including the one presented in this work.

A number of other works have attempted to remedy the computational constraints of the ICWS algorithm in various ways. Most of these works evaluate only one or two scenarios: classification performance, nearest-neighbor precision, or similarity bias. We evaluate all of these scenarios, which will be detailed further in section 4.

Li [10] improved upon ICWS's memory requirements for storage by introducing the 0-bit CWS strategy. The normal CWS sketch contains a sequence of tuples $(z^*, t_{z^*})$, and both values must be equal to consider the pair a match. Li's insight was that if $z^*$ match or don't match between two sketches, it is most probable that $t_{z^*}$ will similarly either match or not. Thus the $t_{z^*}$ can be dropped, while maintain the same fidelity as ICWS in practice. This does not meaningfully improve the runtime of ICWS in most cases, but does reduce the memory needed for storage by half. We refer to this approach as ICWS-0Bit, and it is the inspiration for our improvements. Li's evaluation was done for classification and word bias, two of the three scenarios we evaluate.

Yang et al. [22] looked at leveraging the ICWS-0Bit algorithm. In their work they exploited its structure to create min-hashes over streaming inputs. In this case the total weight coefficients for each histogram input are altered over time. In doing so they develop a related algorithm that can efficiently update a min-hash in $O(K+D)$, but still require $O(KD)$ for initial hash construction. This allows their approach to be orders of magnitude faster to update over re-building the hash over time. We will use a similar approach of simplifying ICWS-0Bit to reach our goals, which is faster initial construction of a min-hash.

---

[1]The value $t_{z^*}$ is technically unbounded in size, but in practice would rarely exceed a 32 bit integer value.

Further work has been done to carefully analyze the ICWS algorithm, and remove redundant steps to reduce the computational requirements while maintaining a mathematically equivalent algorithm. Wu et al. [21] showed that ICWS could be reduced to requiring only four uniform-random samples from a Pseudorandom number generator (PRNG) instead of five, four logarithms/exponentiations, and five floating point multiplications/divisions per feature and hash. Evaluating on classification and nearest neighbor precision (two of the three scenarios we evaluate), they found this reduced runtime by 20–33%, depending on the dataset, and noted the importance of reducing the number of PRNG calls is especially critical as the dataset size increases. In this work, we reduce the cost to just one floating point multiply, require no PRNG calls, and obtain a minimum speedup of over 7x, and up to 28x, dramatically improving upon recent results.

One of the more novel approaches to the WJS problem was presented by Shrivastava [18], who developed a new approach that was not based on the ICWS or CWS algorithms. Their approach's runtime is dependent on the average similarity between points, as well as the largest maximum magnitude per feature. For this reason communication is needed to determine the maximum possible feature value of all possible features before the algorithm can start. This limits their approach's applicability to scenarios with bounded magnitudes and where all features are known up-front. For example, recent work in malware detection wouldn't be able to make use of this approach [17]. When applicable, Shrivastava [18] showed 1500x–6000x speedup for some datasets. Shrivastava also concluded that different approaches to the WJS sketching problem work best for different data sets, and that ICWS should still be preferred when the number of non-zero features is of a similar size as the sketch size $K$. Shrivastava evaluates only estimation bias, which is one of our three scenarios.

## 3 SIMPLIFIED CWS, 0-BITS WITH ONE FLOP

Now that we have reviewed the literature on the Improved Confidence Weighted Sampling algorithm, we show how we can simplify ICWS by extending the reasoning of previous work. In doing so we can construct an implementation of our Simplified ICWS that will require minimal compute time while avoiding expensive PRNG sampling.

### 3.1 Simplifying the ICWS Algorithm

Li [10] showed that the $t_{z^*}$ term in the minhash tuple was not strictly necessary to obtain a high quality approximation of the Weighted Jaccard Similarity. By removing this value from the hash, the size of the hash is reduced by half. Because this uses "zero bits" of the $t_{z^*}$ portion of the hash, it was termed the 0-Bit CWS. This leaves only the selected feature index $z^*$ as the value from the hash itself. This is made possible by the fact that the selected feature index $z^*$ is selected from the $z$ with minimum $a_z$ value, and thus already has information regarding both $t_z$ and the feature's weight.

Our contribution is the realization that if we are using this information, we can relax the procedure given by Ioffe [8] for the ICWS algorithm's consistency property, and thus, the algorithm's implementation. Given a fixed $z, \beta_z$ and $c_z$, the consistency property is shown using the fact that $t_{k^*}$ is a unique integer satisfying the bounds $\frac{\log w(S, z^*)}{r_{z^*}} + \beta_{z^*} - 1 < t_{z^*} \leq \frac{\log w(S, z^*)}{r_{z^*}} + \beta_{z^*}$. Because we do not keep or use the $t_{z^*}$ value, there is no practical need to maintain this bound. Thus we propose to remove the floor function used to compute this value, changing it simply to $t_z \leftarrow \frac{\log w(S, z)}{r_z} + \beta_z$.

This change allows us to propagate several simplifications forward through the ICWS algorithm. First, note that we get

$$\begin{aligned}
y_z &= \exp\left(r_z \left(\frac{\log w(S, z)}{r_z} + \beta_z - \beta_z\right)\right) \\
&= \exp\left(r_z \left(\frac{\log w(S, z)}{r_z}\right)\right) \\
&= \exp\left(\log w(S, z)\right) = w(S, z)
\end{aligned}$$

This allows the immediate removal of one random variable $\beta_z$, and the substituting of $w(S, z)$ for $y_z$ to obtain $a_z = \frac{c_z}{w(S, z) \exp(r_z)}$. With some simple algebra we can re-write this term as $a_z = w(S, z)^{-1} c_z \cdot \exp(-r_z)$. This reduces the mathematical operations from two exponentiations, a logarithm, and four multiplications/divisions to just one exponentiation and two multiplications/divisions. However, four samples from the uniform distribution and an additional four logarithms are still needed to produce $c_z$ and $r_z$. Some minor approximations and reductions can allow us to remove an additional exponentiation and a uniform random sample, at the cost of only one additional multiplication (which is less expensive).

The uniformity property for ICWS was shown by determining that the probability of selecting $a_z$ is equal to $w(S, z)^{-1} \sum_j w(S, j)$ [8]. Given that $c_z \exp(-r_z)$ is a value fixed for all sets $S$, we can show that we maintain this uniformity property. This can be seen by noting that $a_z$ is scaled at a rate of $w(S, z)^{-1}$ by definition. Since we select the minimum value of $a_z$, it corresponds with the maximum value of $y_z$, which as we have just shown, is $w(S, z)$. The Gamma-based terms are independent and so can be marginalized out, leaving the probability of selecting a feature $w(S, i)$ as $w(S, i)/\sum_j w(S, j)$. Thus we maintain the ICWS algorithm's uniformity property without issue.

Because we have altered one of the the values used in the sampling process, our expectation is that this new simplified approach will not match the *exact* behavior of ICWS, where the approaches we reviewed in section 2 do. But since the $t_{z^*}$ was not required, we do expect our new approach to provide similar accuracy and performance in machine learning and information retrieval applications. At first glance, these simplifications may appear to provide little more than what was obtained in prior work that reduced the number of operations in the ICWS algorithm [21]. However, We will show below that our simplification allows for considerable exploitation of the new form of $a_z$, allowing us to dramatically reduce the cost of producing sketches.

## 3.2 Exploiting Simplicity for Efficiency

Now that we have simplified the algorithm, we take critical note that the new $a_z$ definition has the $w(S, z)$ term entirely separated from the functions involving the random Gamma samples $c_z$ and $r_z$. Rather than sample these values as we observe each feature, for each minhash $k$, we can pre-sample a pool of values from the distribution defined by the $c_z \exp(-r_z)$ term. We can use a higher quality PRNG for this step, and the pool need only be sampled once for the entirety of the application. This pooling strategy is only possible because we have removed the coefficient value $w(S, z)^{-1}$ from the computation the distribution of the $a_z$. In ICWS, these values are intertwined and so prevent pre-sampling the resulting distribution.

When iterating over the features, we can select the pre-sampled value from the pool by using a much simpler Linear Congruential Generator (LCG) style PRNG on the feature index $z$ combined with the minhash index $k$. This entire procedure can be found in Algorithm 3, and reduces the ICWS algorithm down to *only one floating point multiply* per feature and hash. Because we have now dramatically reduced the number of FLOPs it is worth noting what other, less costly, operations are being done. This includes one integer multiplication, one integer modulo operation, and a random-access lookup.

---

**Algorithm 3** Simplified CWS (SCWS)

**Require:** An array $T$ of length $|T|$, where $T[i] \sim c_z \exp(-r_z)$, and large primes $p_1$ and $p_2$
1: **procedure** MINHASH(Weighted Set $S$, hash index $k$)
2: 　　$b \leftarrow k p_2$
3: 　　**for all** $z \in S$ **do**
4: 　　　　$\gamma \leftarrow (z p_1 + b) \mod |T|$　　▷ LCG style index selection
5: 　　　　$a_z \leftarrow w(S, z)^{-1} \cdot T[\gamma]$　　▷ The only FLOP needed
6: 　　**end for**
7: 　　$z^* \leftarrow \arg\min_z a_z$
8: 　　**return** $z^*$
9: **end procedure**

---

In our implementation, we choose primes $p_1 = 1073741827$ and $p_2 = 1073741831$. Making these values prime ensures that the modulus operation will result in an index selected uniformly from the pool's size. Because our system has a 32 KB L1 data cache size, we make the pool store 4000 floating point numbers. This ensures that the pool of values will remain in L1 cache, ensuring in turn that the random access lookup will return quickly and keep the procedure from stalling on a memory access. We will see in section 4 that despite this small pool size, we continue to get high quality results with our SCWS algorithm that closely match that of ICWS, while being an order of magnitude faster.

## 4 EXPERIMENTS

We will now describe a number of experiments we performed to validate our new SCWS algorithm, comparing it to the original ICWS algorithm and the ICWS-0Bit algorithm we are inspired by. Because we have made a change to the ICWS algorithm to simplify it as a whole, we will see that our new approach does not closely mimic the original ICWS behavior like ICWS-0Bit does. Instead we gain a significant speed advantage over both ICWS and ICWS-0Bit, while having qualitatively similar results. We will empirically demonstrate that SCWS: 1) continues to return an accurate estimate of the WJS between two points, 2) allows one to efficiently build classifiers using feature hashing, and 3) continues to provide good precision in selecting the true nearest neighbors under the WJS. We conclude our experiments with a test of the pool size to demonstrate that it need not be tuned to any particular problem, and the default size of 4000 is at or past the point of diminishing returns. All code was implemented in Java using the JSAT library [15].

### 4.1 Accurate WJS via Word Similarity

One test for the quality of a CWS scheme was proposed in Li and König [11], where the bias and variance of their approaches were compared using the word document frequencies as the sets. We replicate their approach because it provides a more challenging case for our algorithms due to a heavier tail in the distribution of words. This means the weights for each word in a given document will have a greater variability, and thus better exercise the WJS properties than many common datasets.

To make our protocol reproducible, we specify that we use the 20 News-groups corpus as our collection of documents. We use a simple tokenization on non-alphabetic characters and convert everything to lower-case. Each row corresponds to a document, and each column to a specific word in the corpus. The values in the matrix indicating the number of occurrences of a word in a given document. Our feature vectors for each word are then the columns of the generated data matrix.

For each word-pair, we record the true WJS similarity, and plot the average difference between WJS and each of our CWS algorithms, as we increase the sketch size $k$ from 1 up to 1000. These results can be found in Figure 1. As expected, we see the average difference between WJS and the CWS varieties approach zero as the value of $k$ increases. Each experiment was run 1000 times to obtain a high precision estimate. Because a sketch of size $K$ also contains a sketch of size $K - 1$, we exploit this to provide a point-estimate across all values of $K$ while keeping experimental evaluation time reasonable.[2]

Word pairs were selected to try to cover a diversity of scores and behaviors. In each case, we can see that the ICWS-0Bit algorithm almost perfectly follows that of the original ICWS, with some exceptions, like the "subsidies-settlements" pairing. SCWS clearly does not track the ICWS in exact behavior, but shows the same general characteristics. In some pairings, such as "United-States", all three track together usually, the trackings are close - even if SCWS is slightly more or less accurate, such as "IBM-PC" and "Hong-Kong". Other cases, like "Car-Bike" and "Subsidies-Settlements" show SCWS coming out ahead, though this is not always the case.

Given these results, we can conclude that SCWS is not an unbiased estimate of the ICWS's behavior, as it frequently does not mimic it in the same way the ICWS-0Bit does. But more importantly, we can conclude that SCWS is empirically a high quality estimate of the WJS that retains the fidelity of ICWS's estimates.

---

[2]Generating a new sketch for every value of $K$ would have resulted in an experimental runtime of several months.

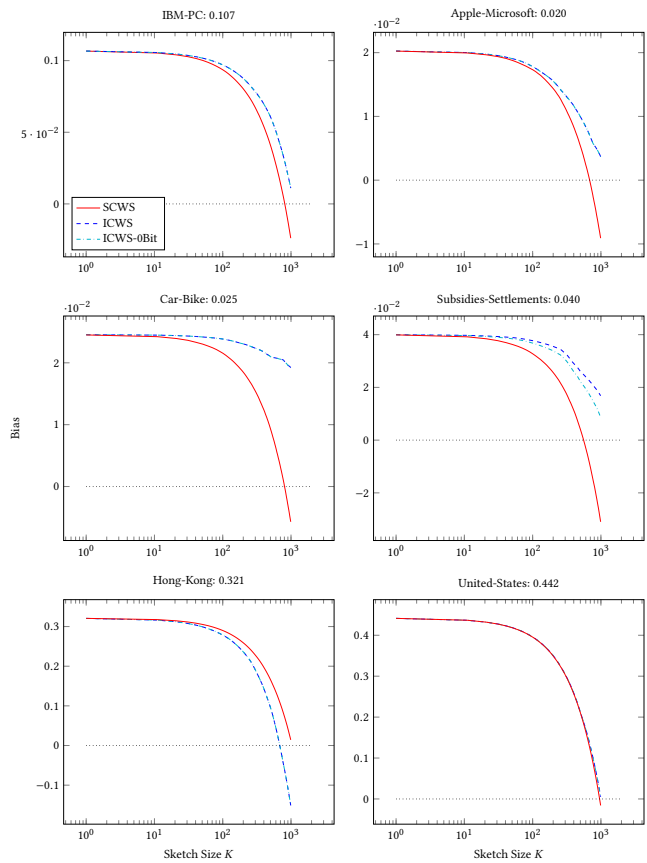

Figure 1: Plots show the difference between each CWS algorithm, and the true WJS. The dotted black line shows the value of zero for a perfect estimate and our new SCWS is in red. Above each figure is the word-pair under test, with the true WJS. The x-axis shows the sketch size $K$, and the y-axis shows the bias of the WJS estimated provided by each CWS compared to the true WJS score. All figures share the same legend.

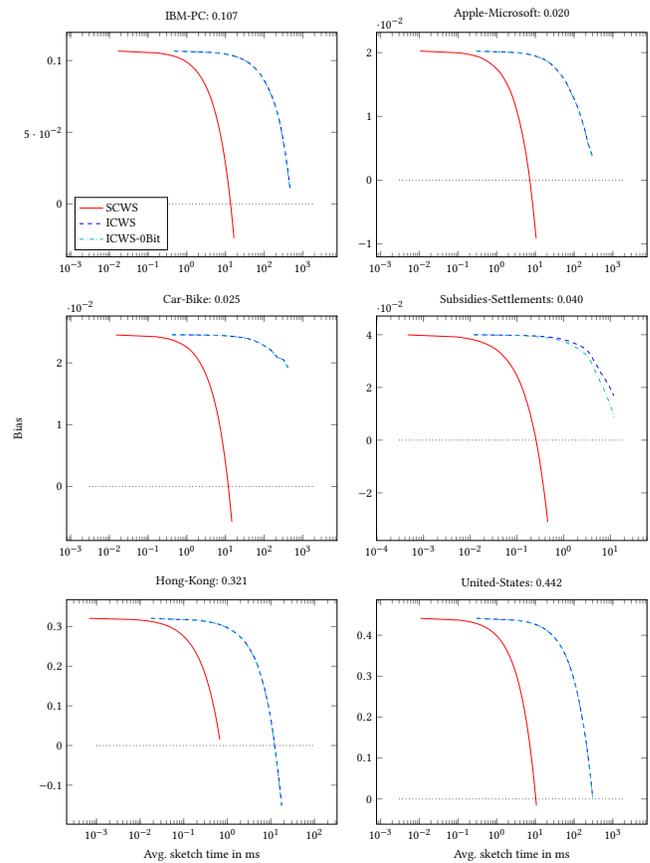

Figure 2: Same as Figure 1, but x-axis replaced with average time to construct the sketch (in milliseconds).

The quality of results is approximately equivalent between all approaches in this test, a theme we will see in other sections as well. The other pertinent issue to the user of any CWS, is how long will it take to obtain some (minimal) level of accuracy? In Figure 1 we plotted the error in the approximated WJS with that of the true WJS as a function of the sketch size. We can instead plot this error with respect to runtime, as shown in Figure 2.

From this view of the data, our SCWS approach uniformly dominates the other curves. The "United-States" word pair is particularly illustrative. Considering just sketch size, all methods were performing at in indistinguishable level. When time becomes a factor, our SCWS sketch construction is up 28.2 times faster to reach zero bias compared to either ICWS variant. This speed advantage is still present when we look at word pairs like "Hong-Kong", where ICWS initially has faster convergence to the true WJS. At sketch size of $K = 1000$, SCWS achieved an absolute error of approximately 0.012, where ICWS achieved this same error rate at only $K = 680$. Even with ICWS having a smaller sketch size, SCWS at $K = 1000$ is 17.6 times faster to construct compared to ICWS at $K = 680$. This strengthens our results, showing that even when SCWS may require a larger sketch size for a particular dataset, the speed advantage can still be an order of magnitude faster.

## 4.2 Learning with SCWS

We now demonstrate that our new SCWS is effective for building binary and multi-class classifiers. Following prior work, we will compare the performance of our approach with that of a linear SVM and a Kernel SVM using the exact WJS as its kernel [10], and show the performance over a range of the regularization penalty $C$ that is common with the SVM. For each of our CWS algorithms, we will use the hashing scheme of Li et al. [12] to create feature vectors from CWS min-hashes. For their approach using 8-bit hashes and $K = 4096$ for our sketch size, and use the features to train a Linear SVM model. The 8-bit hashes with $K = 4096$ combination has been found to give the best classification results with diminished returns when increasing the hash size further [10, 12]. All feature values are re-scaled to the range $[0, 1]$ to avoid any issues with negative weights. Since not all of the datasets we will use have a testing set, we will use the standard training set and estimate generalization with 5-fold cross validation in all cases.

**Table 1: Summary of each dataset used, all sub-sampled so that $N = 20{,}000$. $D$ Indicates the dimension of the dataset, and 'Density' the percentage of non-zero values in the corpus. The right-most column shows how many times faster SCWS was compared to the faster of ICWS and ICWS-0Bit.**

| Dataset | $D$ | Density | Time to Hash (seconds) | | | Speedup |
|---|---|---|---|---|---|---|
| | | | ICWS | ICWS-0Bit | SCWS | |
| a9a | 123 | 11.3 | 186 | 185 | 9 | 19.9 |
| cod-rna | 8 | 99.8 | 106 | 107 | 11 | 9.0 |
| covtype | 54 | 22.1 | 160 | 160 | 12 | 13.1 |
| MNIST | 780 | 19.2 | 1,937 | 1,922 | 86 | 22.1 |
| ijcnn1 | 22 | 59.1 | 173 | 175 | 22 | 7.7 |
| w8a | 300 | 3.9 | 161 | 162 | 10 | 15.4 |
| RCV1 | 47,236 | 0.14 | 661 | 658 | 52 | 12.6 |
| URL | 3,231,961 | 0.004 | 1,516 | 1,491 | 105 | 14.6 |

We perform this evaluation using eight datasets with varying numbers of features and sparsity patterns, all of which are obtained from the LIBSVM website [6]. For each dataset we sub-sample the corpus down to 20,000 samples so that the Kernel SVM will run in a reasonable amount of time. These datasets can be found in Table 1, where we also show how long it took each of the CWS algorithms to produce the feature vectors (running the CWS algorithm, and hashing the returned sketch into the 8-bit feature representation). Of particular note is the right-most column of Table 1, which shows the relative speedup of SCWS compared to the faster of ICWS and ICWS-0Bit. Across all datasets, *SCWS is between 7.7 and 22.1 times faster*.

The results of our approach can be seen in Figure 3, where the performance of our SCWS is comparable to that of ICWS and its 0-bit variant. In some cases SCWS has equal, slightly worse, or slightly superior performance compared to ICWS — depending on the value of $C$ and the dataset under consideration. In most cases we can see the CWS algorithms outperform the linear SVM model, and often equal or outperform the accuracy of the WJS kernelized SVM. Tests were done over a large range of regularization values with $C \in [10^{-3}, 10^2]$. This range proves informative across datasets, such as a9a, in which the CWS approaches matches the best kernel and linear SVMs when given a strong penalty of $10^{-3}$, but drops quickly as $C$ increases. Similarly, all CWS have consistently high performance on the IJCNN corpus, and are beaten by the WJS kernel only for values of $C \geq 10$.

Overall, we argue that our SCWS algorithm shows a high fidelity in approximating the WJS, even if it does not mimic the exact behavior of ICWS in this case. This was predicted by our derivation in section 3, as we noted that the SCWS does make a simplifying change to the original ICWS algorithm. On some datasets, such as RCV1 and MNIST, our new SCWS perfroms better. On datasets like URL and a9a, all CWS approaches produce about the same score. There are also datasets like Covtype and and IJCNN where SCWS performs slightly worse. In each case the standard deviation is shown as a translucent shaded region of the same color, which indicates that the variability in performance is also consistent for each approach on each datasets, even as the value of the regularization parameter $C$ changes.

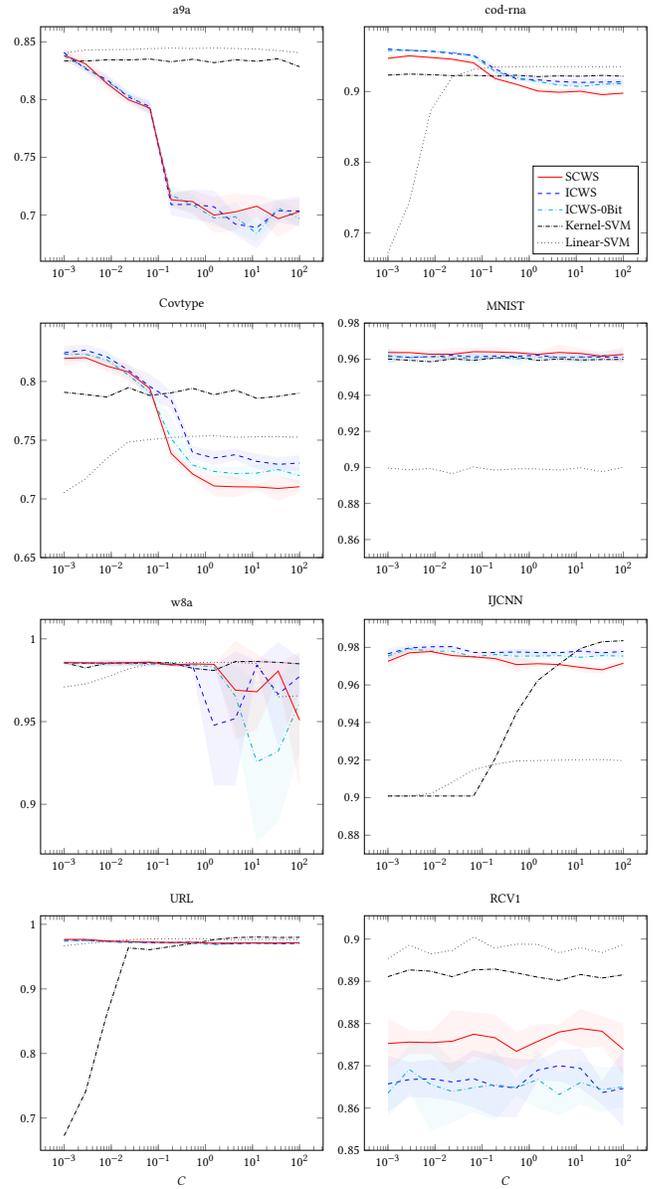

**Figure 3: Performance of linear models built from CWS algorithms, compared to a linear and kernel SVM. The x-axis $C$ shows the regularization parameter's value, and the y-axis shows the accuracy of 5-fold cross validation. Each panel is with respect to a different dataset, which is indicated at the top of each sub-figure. All figures share the same legend. Note each figure has a different scale. Each CWS based method shows has an opaque highlited region indicating $\pm 2\sigma$ (best viewed digitally and in color).**

We note that in the worst case our new SCWS approach was only 7.7 times faster than ICWS in creating the sketch and creating the 8-bit feature vectorization scheme of Li et al. [12]. Comparing ICWS and SCWS in this way is giving an advantage to ICWS, as the vectorization scheme has a similar intrinsic cost regardless of how the sketch was created. As the average number of non-zeros in the dataset increases, so does the relative advantage of our SCWS approach. For instance, on the MNIST dataset, we can see 22 fold speedup of our new SCWS compared to ICWS. This is a dramatic improvement over prior CWS works, which were able to obtain improvements of no more than 40% in execution time [21].

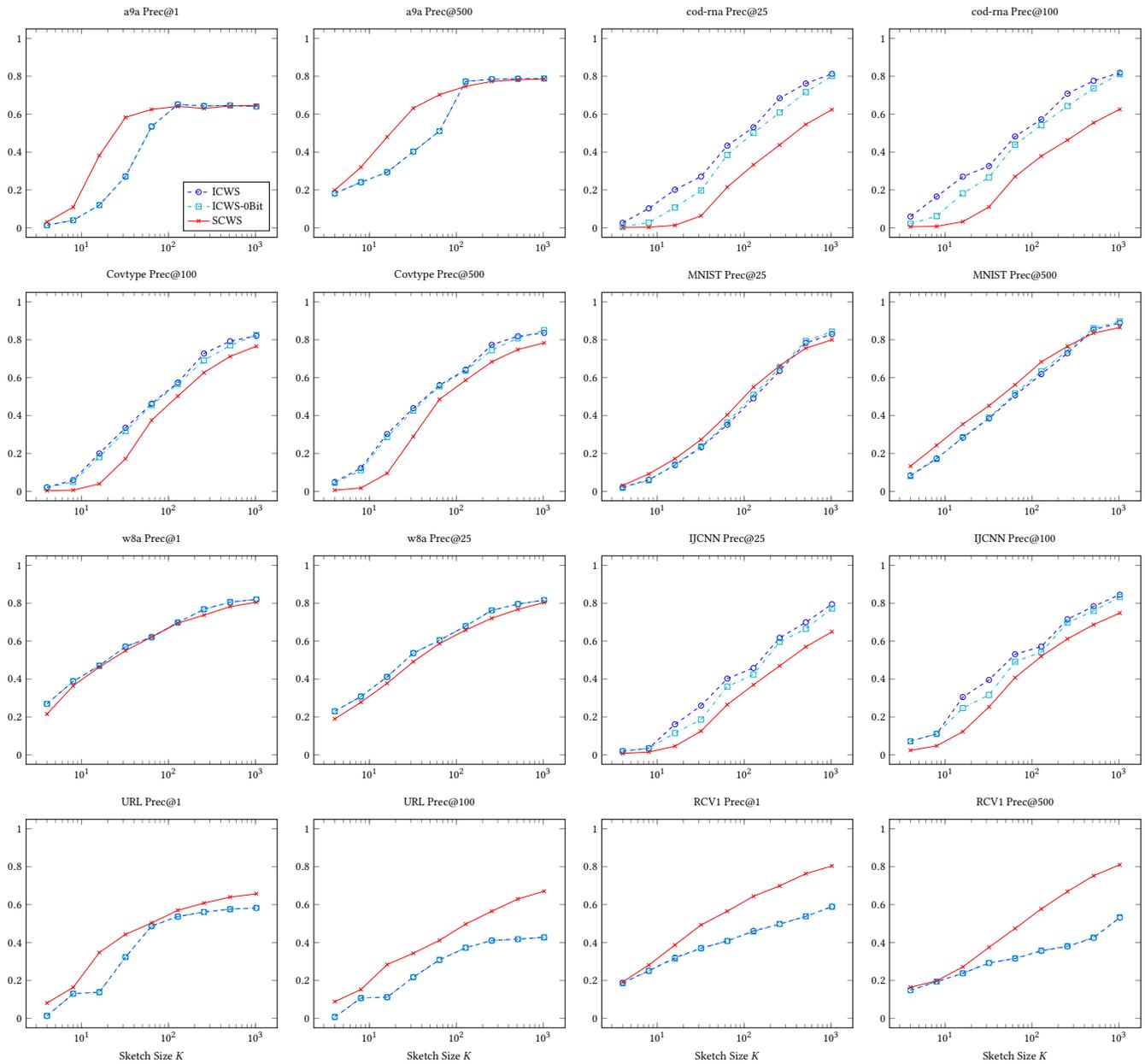

Figure 4: Plots show the precision of selecting the true k-NN according to WJS for each CWS. Above each figure is the dataset under test and neighbor limit. The x-axis shows the sketch size $K$, and the y-axis shows the precision achieved by each CWS. All figures share the same legend and scale.

## 4.3 Nearest Neighbor Precision

In this third test of the effectiveness of our approach, we examine the precision of each CWS algorithm in correctly returning the $k$-nearest neighbors (k-NN) of a point compared to the true WJS. Our experiments follow the protocol used by Wu et al. [21]. We will select a random set of 1000 points from the corpus to use as our query set, and search query points for their nearest neighbors. We will use precision at $\kappa$ as our target metric, measured for multiple sketch sizes and for $\kappa \in \{1, 25, 100, 500\}$. (That is, we measure the precision when returning the single nearest neighbors, the 25 nearest neighbors, etc.) These tests will use the entirety of each corpus, rather than sub-sampling like the previous experiment of subsection 4.2.

We present a selection of these results in Figure 4, showing each dataset twice at two of the four precision levels. This is due to space limitations, but showing the results for two levels allows us to demonstrate that the results are consistent across values of $\kappa$ given a specific dataset. We find that the relative performance of ICWS, ICWS-0 Bit, and SCWS are consistent across precision levels.

In these results we can see the same trend that we saw in the two earlier sets of experiments. SCWS has the same general performance as ICWS, and may have higher, lower, or about equal precision depending on the dataset. In all cases, the SCWS precision is monotonically increasing as the sketch size increases, consistent with it approximating the true WJS.

On the covtype, cod-rna, and IJCNN datasets, the precision of SCWS trails that of ICWS. We also note for these datasets that the 0-Bit approach follows but under-performs compared to the original ICWS algorithm. This may be an indication that while the $t_{z^*}$ term removed from the 0-bit approach (and the random variable $\beta$ tied to it) play an occasional role in improving the quality of results, it is not strictly necessary. The datasets which elicit this gap between ICWS and ICWS-0Bit are the only ones in which SCWS has a meaningful drop in precision.

SCWS outperforms both variants of ICWS on the a9a dataset, and dramatically so on the RCV1 corpus. In the former case, SCWS has superior precision at low sketch sizes, and the approaches converge as the sketch size increases. For the latter case, SCWS starts at the same precision as ICWS for small sketches, but quickly outperforms it as the sketch size increases.

### 4.4 Robustness of Pool Size

The runtime efficiency of our new SCWS method comes from using a pool $T$ of pre-sampled values of the distribution $\sim c_z \exp(-r_z)$, rather than generating new random values manually each time. Values are selected from the pool by a simple indexing strategy to mimic sampling from the true distribution, but requires only one FLOP rather than multiple expensive PRNG generations and multiple FLOPs. An important question regarding the effectiveness of our approach is how big should this pool should be, and what is the pool size's impact on performance?

We can test this by repeating some of our experiments with multiple pool sizes, to look at the change in performance as the size of the pool increases and decreases as other parameters are held constant. We do this for four classification and four precision problems. The results are presented in contour plots, where the y-axis is the size of the pool, and the x-axis shows the regularization parameter $C$ of the SVM or the precision at $\kappa$, for each respective task. The color indicates the accuracy or precision (both on a $[0, 1]$ scale), with respect to the two variables. The goal is that the performance of our SCWS algorithm will be constant with respect to the pool size, once we reach a minimum threshold for the pool's intrinsic size. This would indicate that enlarging the pool of values results in no further improvement on the accuracy of our method. This is also critical to practical deployment of our method. If the performance was overly sensitive to the pool size, it would become a parameter that needs estimation for every new problem, requiring an expensive hyperparameter search. Because we can show that the performance is consistent once a minimum pool size is reached, this minimum size can be used for multiple problems by default, thus avoiding the additional overhead and keeping our approach practical.

As we can see in Figure 5, this behavior bears out in practice. Above the black dashed line, which marks the pool size of 4000 used in all other experiments in this paper, we can see that the color is nearly-constant in the vertical direction. This means for a given regularization penalty $C$ (or precision value $\kappa$), the performance at the 4000 pool size is the same as for any larger pool size, and thus there is no need to waste additional memory increasing the pool of values. Indeed, in these plots it is clear that even a pool of 2 million values would have had no positive impact on accuracy compared to our much smaller, and more practical, 4000 limit.

When one looks at the classification results in the second row (the MNIST, IJCNN, URL, and RCV1 datasets), which all have high accuracy for all values of $C$, we see this reflected in the contour plot. For some datasets, like MNIST, a pool size of only 512 would have been sufficient to get equivalent accuracy results. The a9a, cod-rna, and Covtype datasets had their best performance for small values of $C$, and again, the behavior in the contour plots matches this expectation. For a9a in particular, we obtain the same peak classification performance even when the pool has as few as 32 elements. Thus we can rely on the returned result as being accurate and representative of the CWS's performance. All three of these datasets show slight variability in their performance at higher values of $C$, which is what keeps the contour plots from having a perfectly uniform color. This is observable in the variance of Figure 3 as well, and so is not an issue with our pooling strategy. We note further that the performance in the most critical region of $C$, where the model obtains the best performance, is consistent. The precision tests show a similar pattern of consistency by the 4000 threshold.

Empirically we see that we could use a smaller pool size in many cases, down to about 1,000 entries. This carries a particular importance because this pool size is four times smaller than the hash size itself. When combined with the fact that hundreds of feature values will be indexed into the pool per hash, the pool's size might seem disproportionately small relative to the number of values being accessed from it. That is to say, one's first intuition might be that the pool should be large enough such that values rarely get re-used, but in our case, we will reuse every value in the pool hundreds of times.

Another positive phenomena in these results is that the region of the contour plot with the highest performance (darker-red color as the value approaches 1.0) is also the region of the plot least impacted by decreases to the pool size $|T|$. For example, on a9a SVM plot in the top left corner we see performance stabilize for all values of $\kappa$ at a pool size of about $|T| = 2048$. However, the SVM obtains its best accuracies when the regularize $C \leq 10^{-2}$, and in this region even a pool size of $|T| = 32$ is effective. In future work, any parameter tunning that needs to be done after the application of SCWS could be speed-up by using an adaptive value of $K$.

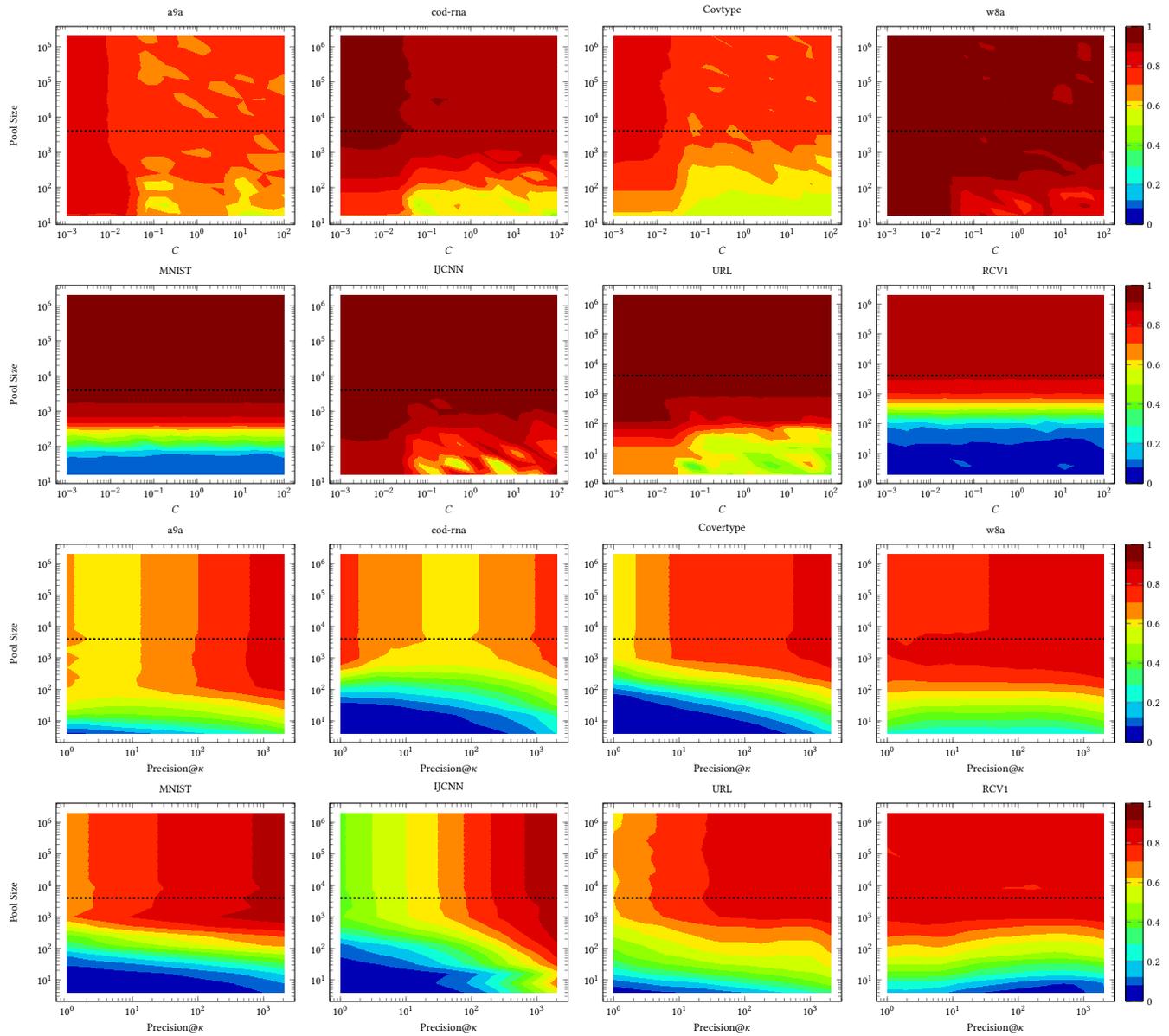

Figure 5: Contour plots showing the impact on pool size with respect to learning performance. The y-axis shows the size of the pool. Color shows the accuracy of the model at each point, and the x-axis is the regularization parameter $C$ or precision rate $\kappa$. First two rows are classification problems, bottom two rows are nearest neighbor retrieval. Each panel is with respect to a different dataset, which is indicated at the top of each sub-figure. All figures share the same legend. The dashed line indicates the standard pool size of 4000 used in all experiments.

The bottom two rows of Figure 5 are the same eight datasets for the information retrieval goal of correctly returning the true $k$-nearest neighbors according to the WJS. The results are further improved in this case where we see almost perfect uniformity of performance with respect to the pool size for every dataset once

$|T| \geq 4,000$. The same trends we discussed above for the classification task (first two rows) are still visible. For example, a9a and MNIST could have obtained equivalent results with a pool size of 128 and 1024 respectively.

It may seem unusual that performance across multiple datasets should be satisfiable with a single pool size $T$. We explain this behavior by interpreting the definition of $a_z = w(S,z)^{-1} c_z \exp(-r_z)$ of SCWS. As was previously discussed, the probability of feature $z$ being selected is directly proportional to the feature value $w(S, z)$, which can be extracted from this definition. The sampled $c_z \exp(-r_z)$ term, which is what the pool contains samples of, then acts as a random perturbation of the ordering — where the probability of perturbation is determined by the distribution, and interacts with the feature value $w(S, z)^{-1}$. Thus the specific value returned by $c_z \exp(-r_z)$ becomes irrelevant; we need only enough distinct values to enable selecting the minimum $a_z$ value in a manner consistent with the true distribution. In this light, we believe it is easy to understand how our approach can work while using such a small pool size.

## 5 CONCLUSION

In this work we have derived a new Simplified CWS algorithm, which enables us to obtain significant speedups up to 22 times faster than the original ICWS algorithm. Through extensive experimental tests we have validated that our new SCWS approach obtains results of equivalent quality. These tests have covered the bias of the approximation on individual points, in ranked retrieval, and classification — ensuring that we have tested a wide gambit of potential use cases. These tests also show that the trick we use to make SCWS fast (i.e. pre-sampling all random values into a finite sized pool) is robust with respect to the pool size. This new approach extends the utility and practicality of the seminal ICWS algorithm. In future work, we hope to develop formal bounds on the approximation error between WJS and our new SCWS.

## ACKNOWLEDGEMENTS

We would like to thank Luke McDowell for his feedback and review of previous drafts of this work.